%% file: main.tex
\documentclass{article}

\PassOptionsToPackage{numbers, compress, sort}{natbib}
\usepackage[main, final]{neurips_2026}

\usepackage[utf8]{inputenc} 
\usepackage[T1]{fontenc}    
\usepackage{hyperref}       
\usepackage{url}            
\usepackage{booktabs}       
\usepackage{amsfonts}       
\usepackage{nicefrac}       
\usepackage{microtype}      
\usepackage{xcolor}         

\usepackage{graphicx}
\usepackage{algorithm}
\usepackage{algorithmic}
\usepackage{amsthm}
\usepackage{amsmath}
\usepackage{dsfont}
\usepackage{thmtools}

\newcommand{\DE}{\overset{\textrm{DE}}{=}}
\def\argmin{\mathop{\rm arg\,min}}
\title{Compressing Collections of Trees with Decision Equivalence}

\author{%
  Hayden McTavish\\
  Department of Computer Science\\
  Duke University\\
  Durham, NC 27705 \\
  \texttt{hayden.mctavish@duke.edu} \\
  \And
  Jon Donnelly\\
  Department of Biostatistics\\
  Johns Hopkins University, Duke University\\
  Baltimore, MD 21205 \\
  \texttt{jdonne18@jh.edu}
  \And
  Margo Seltzer\\
  Department of Computer Science\\
  University of British Columbia\\
  Vancouver, BC V6T 1Z4\\
  \texttt{mseltzer@cs.ubc.ca}
  \And
  Cynthia Rudin\\
  Department of Computer Science\\
  Duke University\\
  Durham, NC 27705\\
  \texttt{cynthia@cs.duke.edu}
}

\begin{document}
\newtheorem{theorem}{Theorem}[section]
\newtheorem{proposition}[theorem]{Proposition}
\newtheorem{lemma}[theorem]{Lemma}
\newtheorem{corollary}[theorem]{Corollary}
\newtheorem{definition}[theorem]{Definition}
\newtheorem{assumption}[theorem]{Assumption}
\newtheorem{remark}[theorem]{Remark}

\maketitle

\input{sections/abstract}

\section{Introduction}
\input{sections/alt-introduction}

\section{Related Work}
\input{sections/related-work}

\section{Preliminaries}\label{sec:preliminaries}
\input{sections/method}

\section{Experiments}\label{sec:experiments}
\input{sections/experiments}

\section{Conclusion}
\input{sections/conclusion}

\section{Acknowledgments} This work relates to Department of Navy award N000142612253 issued by the Office of Naval Research. We acknowledge funding from the National Institutes of Health (NIH) under grant R01-DA054994. Additionally, this material is based upon work supported by the National Science
Foundation Graduate Research Fellowship under Grant No. DGE 2139754. We acknowledge the support of the Natural Sciences and Engineering Research Council of Canada (NSERC). Nous remercions le Conseil de recherches en sciences naturelles et en génie du Canada (CRSNG) de son soutien. We would also like to thank Elif Arslan for comments helpful to implementing the STreeD package modification to support conservative partitioning (Algorithm \ref{alg:streed-change}), as well as Zachery Boner and Zakk Heile for comments on earlier iterations of the work.

\bibliography{main}
\bibliographystyle{icml2025}


\appendix

\input{sections/appendix/algorithms}
\input{sections/appendix/theory}
\input{sections/appendix/additional_results}
\input{sections/appendix/scoping.tex}


\newpage

\end{document}

%% file: sections/abstract.tex
\begin{abstract}
Decision trees with binary leaves are fundamental building blocks in classical machine learning. While individual trees can be used as strong interpretable classifiers, they frequently serve as members of large collections instead: primarily as components of powerful ensembles, but also as one of the few model classes for which the entire set of near-optimal models (the Rashomon set) can be analyzed. Both types of collections are often more complex than necessary, because there are many decision tree representations for the same decision boundary, imposing unnecessary computational costs in ensembles and complicating theoretical work on Rashomon sets. We propose methods to reduce collections of trees to a subset representing each unique decision boundary in the collection, with each tree in its sparsest equivalent form. In so doing, we improve the scalability of removing provably redundant classifiers in a Rashomon set. We also compress many shallow Adaboost ensembles by >15\% of their total number of leaves, while exactly preserving the models' decisions and probability estimates on any possible observation. These methods allow practitioners to run more efficient computations over compressed collections of trees and remove arbitrary complexity when analyzing and understanding them.
\end{abstract}

%% file: sections/alt-introduction.tex
Collections of decision tree classifiers are core to classical machine learning. A number of popular tabular classifiers are ensembles of trees \citep{freund1997decision, breiman2001random}. These ensembles consistently rank near the top of tabular benchmarks \citep{fernandez2014we, grinsztajn2022tree}. They perform well on accuracy as well as metrics of calibration and ranking, such as ROC-AUC \cite{caruana2006getting, ertekin2011equivalence}. Even when recent work focuses on finding optimal single, interpretable trees as the final classifier instead of an ensemble \citep{rudin2022interpretable, costa2023recent, gosdt, dl85, streed, chaoukibranches}, collections of trees become relevant from a different perspective. 
Simpler models are most likely to be competitive to black box models when there are many near-optimal models for a dataset \citep{semenova2022existence, semenova2023path, boner2024using}. In such settings, it can be of great benefit to materialize and study this set of near-optimal models, called the Rashomon set \citep{breiman2001random, RudinEtAlAmazing2024}. One of the few model classes for which the Rashomon set can be materialized and studied exactly is the model class of decision trees \citep{xin2022exploring, arslan2025sorted}. 

It would be ideal to remove redundancies in these collections before using them in downstream tasks. For ensemble classifiers, reducing ensemble size corresponds to improved memory and energy footprint \cite{buschjager2023joint}. Compressed ensembles become easier to explain \citep{bassan2025makes} and to verify for properties such as fairness and robustness \cite{kantchelian2016evasion, devos2023detecting}. For Rashomon sets, a key question relates to disentangling which trees can make distinct predictions on future data and which trees have distinct structure but identical predictions \citep{semenova2023path, nguyen2025unique}. Frequently, Rashomon sets do contain a large proportion of redundant trees \citep{mctavish2025leveraging}, and their removal improves downstream applications such as variable importance \cite{donnelly2023rashomon}, but removing these trees scalably remains a challenge.

Collections of decision tree classifiers often exhibit redundancy due to \textit{decision equivalence} \cite{cockett1987decision, zantema1998decision}; two trees can make the same classification decisions for any possible sample (even one that is out of distribution) despite having distinct structures. This can cause collections to be redundant in two ways. Collections may contain trees exhibiting decision equivalence to each other, which is a frequent occurrence in Rashomon sets \cite{mctavish2025leveraging}. Moreover, individual trees that are trained greedily, such as the trees in many ensembles, can be losslessly reduced to a sparser, decision equivalent tree that corresponds to the exact same function.

We reduce collections of decision trees by removing decision equivalent trees and simplifying each remaining tree to sparsest equivalent form. To remove decision equivalent trees, we present a polynomial-time algorithm that builds a dataset such that every decision equivalence class in the collection has a different classification pattern. To simplify trees to their sparsest form, we create a reduction to a classification task with missing values based on each leaf of the tree and adapt a state-of-the-art optimal decision tree solver framework, STreeD \citep{streed}, to the task.

This reduced form lets us solve two key problems. First, we can compress classical ensembles, such as Adaboost \cite{freund1997decision}, while exactly preserving the ensemble's ROC-AUC and predicted probabilities. While several existing approaches can compress ensembles while guaranteeing faithful classification decisions \cite{vidal2020born, emine2025free}, they do not guarantee preservation of predicted probabilities or ROC-AUC. Second, we improve the scalability of algorithms to remove decision equivalent forms from a Rashomon set and generalize these checks to identify provable redundancies owing to discretization of continuous features.

 Our core contributions are as follows. 
 
 \begin{enumerate}
     \item We provide tools to scalably filter the occurrence of decision equivalence in collections of trees. We demonstrate lossless compression of decision tree Rashomon sets several orders of magnitude faster than prior work (with improved asymptotic complexity) and handle equivalence arising from multiple splits on the same underlying continuous or categorical feature.
     \item We provide an algorithm to simplify a decision tree to its sparsest equivalent form, demonstrating practically feasible compression of shallow Adaboost ensembles while exactly preserving their predictions and calibration.
 \end{enumerate}

%% file: sections/related-work.tex
\textbf{Decision Equivalence.} The fact that many decision tree expressions map to the same decision boundary has a long history of exploration. \citet{cockett1987decision}  outlines an approach to compute equivalence efficiently and provides a complete set of operations to manipulate between decision equivalent forms. \citet{cockett1990decision} then recommend using a subset of these operations to scalably simplify decision trees (though not with guarantees of finding their sparsest equivalent form). \citet{zantema1998decision} formalize an algorithm for evaluating decision equivalence between a pair of trees with time quadratic in the number of leaves. \citet{zantema1999finding} go on to prove that finding the sparsest form of a given tree is NP-hard. Decision equivalence has also been explored more recently by various other terms, often in the context of collections of trees. Several such terms are faithful \citep{vidal2020born}, predictively equivalent \citep{mctavish2025leveraging}, and functionally identical \cite{emine2025free}. Decision equivalence has a number of convenient decompositions in relation to the leaves of trees \cite{cockett1987decision, cockett1987discrete, cockett1990decision}. These leaves correspond to an implicant \citep{quine1952problem}, a weak abductive explanation \citep{ignatiev2019abduction, izza2022tackling}, and a sufficient reason \citep{ribeiro2018anchors} for that tree's prediction. In this work we refer to the subspace that is routed to a particular leaf as a leaf subregion.

\textbf{Decision Equivalence for Tree Ensembles.} In this work, we focus on decision equivalence within ensembles of decision tree classifiers with binary-valued leaves, primarily focusing on Adaboost \cite{freund1997decision} (the method also applies to random forests \cite{breiman1984classification})\footnote{The method applies to Random Forests as described in the original paper, although scikit-learn's implementation \citep{scikit-learn} uses leaf probabilities instead of binary-valued leaves. Conversely, the original Adaboost can use binary-valued leaves or leaf probabilities, but scikit-learn only supports the former; we focus on scikit-learn's Adaboost implementation in our experiments.}. Here the predicted probabilities are a function of a (weighted) sum of the predictions of each individual tree. 
Several works find sparser forms of tree-based ensembles that are guaranteed to make identical classifications for any sample. \citet{vidal2020born} present an algorithm to find a single decision tree of minimal size with identical predictions to a random forest of trees, using a branch and bound search over the intersections of leaves in the forest. \citet{emine2025free} convert from an additive ensemble to a sparser decision equivalent ensemble by learning a sparse set of weights over the original set of trees. By contrast, our simplification of ensembles preserves both the predicted classes of the ensembles for any possible sample and the predicted probabilities. 

\textbf{Rashomon Sets} The Rashomon effect, introduced in machine learning by \citet{breiman2001statistical}, refers to the phenomenon whereby there are many nearly equivalently good classifiers for a given machine learning task. The full set of these nearly-equivalent classifiers for a given model class, the Rashomon set, is of interest for a range of applications, including robustness \citep{hsu2025double}, variable importance \cite{donnelly2023rashomon}, understanding when simple models will approach the performance of complex models \citep{semenova2022existence, semenova2023path, boner2024using}, and allowing customization and control to domain experts and practitioners \citep{RudinEtAlAmazing2024}. 
Decision trees are one of the few model classes for which an exact Rashomon set with respect to regularized empirical risk can be computed relatively tractably \cite{xin2022exploring, arslan2025sorted}. However, some applications, such as variable importance, can be confounded by the existence of decision equivalent trees in the Rashomon set \cite{mctavish2025leveraging}. This confound has motivated researchers to try a range of adaptations, such as clustering trees by predictive behavior on observed data \citep{nguyen26a} or constraining to a subspace of tree-based partitions where decision equivalence is easier to identify \citep{venkateswaran2024robustly}.

Two approaches have been proposed to identify decision equivalence in Rashomon sets of decision trees. 
\citet{mctavish2025leveraging} discuss ways that accounting for decision equivalence can improve various downstream classification tasks using trees. Several of the paper's investigations relate to compressing Rashomon sets to preserve only one instance of each decision equivalence class. The authors map each tree to a simplified Quine-McCluskey form \citep{quine1952problem, mccluskey1956minimization}, and then do a linear pass through the Rashomon set to remove trees with identical forms. This scales linearly in the size of the set, but each Quine-McCluskey reduction is NP-hard in the size of the trees, so this is only suitable for the Rashomon sets of shallow trees studied in that work. \citet{marquessilva2025efficientpredictive} propose using pairwise decision equivalence operations for this task, to allow scaling to larger individual trees. When we are dealing with a Rashomon set, for which the size itself can scale exponentially with the allowed complexity of each component tree, running equivalence checks between every pair in the set can be a substantial computational drawback; but for any collection with complex component trees, a Quine-McCluskey reduction can also be prohibitively expensive. Our work explores scalability improvements for this task that are more efficient per-tree than the approach in \cite{mctavish2025leveraging} and more efficient over the entire set than the approach in \cite{marquessilva2025efficientpredictive}. 

\textbf{Optimal Decision Tree Classifiers.} A number of works globally optimize a decision tree's classification accuracy with a penalty or constraint on the complexity of the corresponding tree. While this has taken many avenues, including mixed integer programming and SAT solvers \cite{bertsimas2017optimal, verwer2019learning}, more recent approaches have found success with branch-and-bound dynamic programming approaches \cite{dl85, gosdt, murtree, streed, chaoukibranches}. In this work, we discuss a minor modification to branch-and-bound search strategies for optimal trees to handle missing data, and use this to allow a reduction from finding the sparsest form of a tree to finding an optimal tree on a simple dataset based on the leaves of the original tree.

%% file: sections/method.tex
Define feature space $\mathcal X \subseteq \mathbb{R}^p$, where each of the $p$ features is continuous, categorical, or binary. We denote a training dataset with $n$ samples as $\mathcal D = \{(x_i, y_i)\}_{i=1}^n$, where each $x_i \in \mathcal X$ and each $y_i \in \{0, 1\}$. 
Define a boolean query $q = (j,t)$ corresponding to the function $\mathds{1}\{v_{j} \leq t\}$ (where $\mathds{1}$ is the indicator function) for an input vector $v \in \mathcal X$. Let $queries(T)$ be the list of all unique boolean queries $(j, t)$ in a tree $T$. 
A binary decision tree classifies samples by making a series of boolean queries over individual features, routing samples to a left or right subtree based on the query, until reaching a leaf corresponding to a prediction of $0$ or $1$. Define the class of all binary trees as $\mathcal T$. Denote $T(x)$ as the prediction of tree $T \in \mathcal T$ for sample $x$.
Let ${depth}(T)$ be the depth of the tree (the maximum number of queries in any path of the tree), and let the class of trees of depth at most $d$ be $\mathcal{T}_d = \{T \in \mathcal{T} \mid {depth}(T) \leq d\}$. 
Denote the size of the tree $T$ by $|T|$, and let the size correspond to the number of leaves in the tree.

The leaves of decision tree T, denoted as $leaves(T)$, correspond to a partition of the feature space $\mathcal{X}$ based on the queries to reach each leaf. 
Let $\mathcal I(\mathbb{R})$ denote the set of all intervals in $\mathbb{R}$. A leaf subregion $S(\ell) \in \mathcal I(\mathbb{R})^p$ corresponds to the subset of the feature space $\mathcal X$ that routes to leaf $\ell$. 
For any sample $x_i$, the sample reaches leaf $\ell$ if and only if $x_{ij} \in S(\ell)_j$ for every feature $j \in [1, \ldots, p]$. Note that for any features $j'$ not along the root-to-leaf path for leaf $\ell$, $S(\ell)_{j'}$ is the open interval $(- \infty , \infty)$.
The prediction for any samples reaching leaf $\ell \in leaves(T)$ is denoted $\hat{y}_\ell$. 

We define decision equivalence, denoted by $\overset{\textrm{DE}}{=}$, as a guaranteed correspondence in the predictions of two trees for any possible sample: 

\begin{definition}[Decision Equivalence]\label{def:DE-basic}
    $T_1 \DE T_2 \iff \forall x \in \mathcal{X}, T_1(x) = T_2(x) $,
\end{definition}

where we note that decision equivalence is defined independently of the observed data $\mathcal{D}$.

Decision equivalence allows us to define a sparsest form for a tree (with a constraint on depth to keep our optimization task feasible and to prevent increases to the worst case inference time of a tree).
\begin{definition}[Sparsest Form]\label{def:sparsest-form}
Let $T$ be some decision tree. For some depth $d$, denote the set of all sparsest forms of tree $T$ in $\mathcal T_d$ as 
    $$\textsc{Sp}(T, d) := \argmin_{\{T' \in \mathcal{T}_{d} \mid  T' \DE T\}} |T'|$$
\end{definition}

Given some collection $C_d \subseteq \mathcal T_d$, we can define its reduced form as having only one tree, in sparsest form, for each decision equivalence class in $C_d$: 
\begin{definition}[Reduced Form]\label{def:reduced-form}
$$\textsc{Reduced}(C_d) := \argmin_{C'_d \subseteq \mathcal{T}_d} \{|C'_d| \textrm{ s.t. } \forall T \in C_d, \exists T' \in C'_d, T' \in \textsc{Sp}(T, d)\}.$$
\end{definition}

\paragraph{Completions.} To represent how queries are evaluated on a leaf subregion, it is convenient to work in a feature space of booleans that may be unknown.

Let $M \subseteq \{0, 1, NA\}^q$ be the space of boolean features with potentially unknown (NA) values. A dataset in this space of size $n'$ is then $\mathcal D_{missing} = \{(m_{i'}, y_{i'})\}_{i'=1}^{n'}$ with feature vectors $m_{i'} \in M$. A $\textsc{Completion}$ of $m_{i'}$ is a possible replacement of each $NA$ value in the feature vector with $0$ or $1$. Define the set of all such completions as: 
$$\textsc{Completions}(m_{i'}) := \{z_{i'} \in \{0, 1\}^q \mid \forall j \in [1, \ldots, q], z_{i'j} = m_{i'j} \vee m_{i'j} = NA \}.$$
Similarly, define a completion of a whole dataset as any possible completion of each $m_i$. So the set of all possible completions of a dataset $D_{missing}$ is: 
$$\textsc{Completions}(\mathcal D_{missing}) := \{\{(z_{i'}, y_{i'})\}_{i'=1}^{n'} \mid \forall i', z_{i'} \in \textsc{Completions}(m_{i'})\}.$$

\paragraph{Rashomon Sets.} A Rashomon set for some observed dataset $\mathcal D$ describes the set of all models whose objective falls within some specified tolerance $\varepsilon$ of the optimum. For the model class of decision trees, Rashomon sets are often defined with respect to the following objective \citep{xin2022exploring, arslan2025sorted}: 
\begin{align}\label{eq:optimal-tree-objective}
    \textrm{Obj}(T, \mathcal D, \lambda) = \frac{1}{n}\sum_{i=1}^n \mathds{1}\{T(x_i) \neq y_i\} + \lambda |T|,
\end{align}
for some leaf penalty $\lambda$. The Rashomon set for objective tolerance $\varepsilon$ and depth constraint $d$ is then: 
\begin{align}\label{eq:rset}
    \textrm{Rset}_\varepsilon(\mathcal D, \lambda, d) = \{T \in \mathcal T_d  \mid  \textrm{Obj}(T, \mathcal D, \lambda) \leq \min_{T' \in T_d} \textrm{Obj}(T', \mathcal D, \lambda) + \varepsilon\}.
\end{align}

\section{Methods}

We now outline our methods that find reduced forms for collections of trees (Definition \ref{def:reduced-form}). This decomposes into two tasks: preserving only one tree from each set of decision equivalent trees (Section \ref{sec:pairwise}) and reducing each tree to its sparsest equivalent form (Section \ref{sec:sparsest}).

\subsection{Identifying Decision Equivalence in a Collection}\label{sec:pairwise}

Given a collection of trees $C_d$, we would like to preserve only one tree per decision equivalence class: 

\begin{align}\label{eq:collection}
    \argmin_{C'_d \subseteq \mathcal{T}_d} \{|C'_d| \textrm{ s.t. } \forall T \in C_d, \exists T' \in C'_d, T' \DE T\}
\end{align}

To accomplish this, we iterate through a collection of trees, adding trees one at a time to a new collection, and skipping trees for which a decision equivalent tree already exists. 
To minimize the total number of decision equivalence comparisons we need to make, we maintain a dataset of samples for which no two trees in the current collection have exactly the same prediction vector. 
This ensures that each new tree we add requires at most one decision equivalence check.

Algorithm \ref{alg:collection} presents the approach. After initializing a hashmap \textsc{Pred\_Map} to store the trees we will keep in our compressed collection, and a list of samples $X^{(\textrm{diff})}$, we iterate through the collection of trees ($C_d$), maintaining the invariant that we can use the predictions on $X^{(\textrm{diff})}$ to separate all the decision equivalence classes we have seen so far. For each tree in $C_d$, we check if its predictions on $X^{(\textrm{diff})}$ correspond to a tree already in \textsc{Pred\_Map} (lines 6-9). 
If we do have a collision for all samples in $X^{(\textrm{diff})}$, we check if the two trees are decision equivalent (line 11). 
If there is a data point for which the trees differ, then we update the colliding tree and the new tree's key, add both to $C'$ (lines 12-16), and update $X^{(\textrm{diff})}$. If the trees were decision equivalent, then we can safely leave $T$ out of the compressed set and only keep $T'$ (line 16).
If a tree $T$ never hits a collision in $\textsc{Pred\_Map}$, then it is guaranteed not to be equivalent to any tree currently in \textsc{Pred\_Map}, so we can add it without modifying $X^{(\textrm{diff})}$ (line 19).

\begin{algorithm}[h]
\caption{\textsc{SimplifyCollection}($C_d$)}
\begin{algorithmic}[1]\label{alg:collection}
\item[] {\bfseries Input:}  Collection of trees $C_d$.
\item[] {\bfseries Output:} A compressed version of $C_d$ satisfying Equation (\ref{eq:collection}); only one member of each decision equivalence class is preserved.
\STATE $X^{(\textrm{diff})} \gets []$ \COMMENT{dataset to distinguish between trees in our collection}
\STATE $\textsc{Pred\_Map} \gets \{\}$ \COMMENT{Map predictions on $X^{(\textrm{diff})}$ to a tree in the compressed set}
\item[] (Invariant: a decision equivalence class's index into \textsc{Pred\_Map} is always a prefix of predictions on $X^{(\textrm{diff})}$ that is unique from all other equivalence classes in \textsc{Pred\_Map}.)
\FOR{$T \in C_d$}
\item[] \COMMENT{Start by finding prefix of predictions on $X^{(\textrm{diff})}$ in \textsc{Pred\_Map}, if one exists: }
\IF{$\exists k \leq |X^{(\textrm{diff})}| \textrm{ s.t. } [T(x_i) \text{ for } x_i \in X^{(\textrm{diff})}_{1:k}] \in \textsc{Pred\_Map}$}
\STATE \textrm{potential\_collision} $\gets 
[T(x_i) \text{ for } x_i \in X^{(\textrm{diff})}_{1:k}]$
\STATE T' $\gets \textsc{Pred\_Map}.pop(\textrm{potential\_collision})$ \COMMENT{remove collision from \textsc{Pred\_Map}}
\IF{$\exists k' \textrm{ s.t. } |X^{(\textrm{diff})}| \geq k' > k,  T(x^{(\textrm{diff})}_{k'}) \neq  T'(x^{(\textrm{diff})}_{k'})$} 
\item[] \COMMENT{{If there are multiple $k'$, any work; our approach uses the smallest}}
\STATE $\textsc{Pred\_Map}[T (x_i) \textrm{ for } x_i \in X^{(\textrm{diff})}_{1:k'}] = T$
\STATE $\textsc{Pred\_Map}[T' (x_i) \textrm{ for } x_i \in X^{(\textrm{diff})}_{1:k'}] = T'$
\ELSE
\STATE x $\gets \textsc{FindNonEquivalence}(T,T')$ \COMMENT{Algorithm \ref{alg:de}; returns an $x \textrm{ s.t. } T(x) \neq T'(x)$}
\IF{x $\neq $ None} 
\STATE $X^{(\textrm{diff})} \gets X^{(\textrm{diff})} + $[x] \COMMENT{Augment $X^{(\textrm{diff})}$ to maintain invariant}
\STATE $\textsc{Pred\_Map}[T(x_i) \textrm{ for } x_i \in X^{(\textrm{diff})}] = T$ \COMMENT{collision resolved and $T \not \DE T'$, so add $T$}
\ENDIF
\STATE $\textsc{Pred\_Map}[[T'(x_i) \textrm{ for } x_i \in X^{(\textrm{diff})}]] = T'$ \COMMENT{add $T'$ back in since we popped it earlier}
\ENDIF
\ELSE
\STATE $\textsc{Pred\_Map}[T(x_i) \textrm{ for } x_i \in X^{(\textrm{diff})}] \gets T$ \COMMENT{no collision in \textsc{Pred\_Map}}
\ENDIF
\ENDFOR
\STATE Return $\textsc{Pred\_Map}.\textrm{values()}$
\end{algorithmic}
\end{algorithm}

Algorithm \ref{alg:de} in the appendix presents the pairwise equivalence subroutine we use. If the two provided trees are not equivalent, the algorithm returns a data point for which the trees' predictions differ, which helps in constructing  $X^{(\textrm{diff})}$. If the two trees are equivalent, the algorithm returns None. We defer this algorithm to the appendix, because there exist algorithms with similar scalability \citep{cockett1987decision, zantema1998decision}. The key difference in our algorithm is that we can handle equivalences where boolean splits relate to the same underlying continuous feature, at no additional asymptotic cost. Section \ref{sec:experiments} demonstrates that, for several Rashomon sets trained on real-world data, these continuous feature dependencies account for a substantial proportion of all decision equivalence.

Proposition \ref{thm:collection}, proven in Appendix \ref{app:proofs}, shows that the algorithm is correct and provides an upper bound on its runtime. If the 
collection size is sufficiently large, then a simple upper bound for the runtime becomes $\mathcal O (d|C_d|^2)$. The approach gives a substantial improvement on the asymptotics of existing approaches. 
It removes the need for the NP-hard subroutine used by \citet{mctavish2025leveraging}. It also avoids the $\mathcal O(|C_d|^2)$ pairwise checks needed for a na\"ive compression, where each pairwise check would require time quadratic in the leaves of the two compared trees.

\begin{restatable}{proposition}{collectioncorrectness}\label{thm:collection}
     Let the original collection be denoted $C_d$, and let the domain of the trees have dimensionality $p$. Let the output of Algorithm \ref{alg:collection} applied to $C_d$ be $C'_d$. Then, $C'_d$ satisfies equation (\ref{eq:collection}), and the algorithm has runtime in 
     $$\mathcal O \left(|C_d| \left(|C'_d|d + \max_{T \in C_d, T' \in C'_d} |T||T'| + p\right)\right).$$
\end{restatable}

\textbf{Reducing Rashomon sets.} Decision tree Rashomon sets can also be enumerated in sorted order of objective \cite{arslan2025sorted}. If we traverse a collection in this sorted order and preserve only the first tree encountered for each decision equivalence class, then Remark \ref{thm:rashomon-reduction} guarantees the result is in Reduced Form (Definition \ref{def:reduced-form}), without the need to run a separate sparsest form solver.

\begin{restatable}{remark}{collectionRashomon}\label{thm:rashomon-reduction}
    Let $R = \textrm{Rset}_\epsilon(\mathcal D, \lambda, d)$ be a Rashomon set for some dataset $\mathcal D$, slack $\epsilon$, maximum depth $d$, and leaf penalty $\lambda > 0$. Then its reduced form corresponds to preserving only one tree of lowest objective within each decision equivalence class. 
    $$\argmin_{C'_d \subseteq R} \{|C'_d| \textrm{ s.t. } \forall T \in R, \exists T' \in C'_d, T' \DE T, \textrm{Obj}(T', \mathcal D, \lambda) \leq \textrm{Obj}(T, \mathcal D, \lambda)\} \in \textsc{Reduced}(R).$$
\end{restatable}

\subsection{Reducing Trees to Sparsest Form}\label{sec:sparsest}

Given a decision tree $T_\textrm{original}$, we would like to find a decision equivalent tree of minimal size (that is, a member of $\textsc{SP}(T_\textrm{original}, d)$ from Definition \ref{def:sparsest-form}). 
While this problem is NP-hard \citep{zantema1998decision}, we can leverage recent advances in optimal sparse decision tree solvers 
to find sparsest forms for small trees tractably. Ordinarily, an optimal sparse decision tree solver \citep{gosdt, streed} would find a minimizer for objective (\ref{eq:optimal-tree-objective}) from Section \ref{sec:preliminaries}. Finding the sparsest form of a tree is a distinct task: we want a tree that predicts exactly the same as the original tree on a potentially infinite set of points (any $x \in \mathcal X$), while being as sparse as possible. To accomplish this, we introduce two modifications: we show how to create a special dataset that partitions the space $\mathcal X$ according to the leaves of $T_\textrm{original}$, and we make a small adjustment to the partitioning behavior of an optimal solver to handle this dataset appropriately. 

\textbf{Leaf dataset.} In Algorithm \ref{alg:make-dataset}, we outline a way to construct a compact dataset of size $|T_\textrm{original}| \times |\textrm{queries($T_\textrm{original}$)}|$, such that each training sample corresponds to a leaf in tree $T_\textrm{original}$. Each column corresponds to one of the queries in tree $T_\textrm{original}$, and a row of the dataset evaluates to 1, 0, or NA for each query depending on whether the query holds for all possible samples routed to the leaf corresponding to that row. A tree that correctly classifies all possible completions of this dataset corresponds to a decision equivalent tree (see Lemma \ref{remark:leaf-dataset}, proven in the appendix).

\begin{restatable}{lemma}{LeafDataset}\label{remark:leaf-dataset}
    Let $\mathcal D_{missing} = \{(m_{i'}, y_{i'})\}_{i'=1}^{|T'|}$ be the output of Algorithm \ref{alg:make-dataset} for tree $T'$. 
    If the worst-case classification error of tree $T$ across any completion is $0$, then $T \DE T'$. If no path in $T$ or $T'$ is impossible (contains a query with value already known from earlier queries along that path), and queries are all of the form $\mathds{1}\{x_{ij} \leq t\}$, then this becomes bidirectional: 
    $$T \DE T' \iff \left(\sum_{i'=1}^{|T'|} \max_{x_{i'} \in Completions(m_{i'})} \mathds{1}\{T(x_{i'}) \neq y_{i'}\}\right) = 0$$
\end{restatable} 
\begin{algorithm}[t]
\caption{\textsc{LeafDataset}(T)}
\begin{algorithmic}\label{alg:make-dataset}
\STATE {\bfseries Input:}  Decision tree $T$ 
\STATE {\bfseries Output:} Dataset $X_T \in \{0, 1, \textrm{NA}\}^{|T|\times|queries(T)|}$, $y_T \in \{0, 1\}^{|T|}$
for which perfect classification on all possible completions of this dataset is necessary and sufficient for a tree to be decision equivalent to $T$.
\STATE $X_T \gets []$
\STATE $y_T \gets []$
\FOR{$\ell \in leaves(T)$}
\STATE $x_T \gets []$
\FOR{$k \in |queries(T)|$}
\STATE $j, t \gets queries(T)[k]$ \COMMENT{Collect $j, t$ from query of form $\mathds{1}[x_{ij} \leq t]$}
\IF{$S(\ell)_j \subseteq (-\infty, t] $} 
\STATE $x_T.\textrm{append}(1)$ \COMMENT{we know $\forall x_i \in S, x_{ij} \leq t$, so append $1$}
\ELSIF{$S(\ell)_j \subseteq (t, \infty)$}
\STATE $x_T.\textrm{append}(0)$ \COMMENT{we know $\forall x_i \in S, x_{ij} > t$, so append $0$}
\ELSE
\STATE $x_T.\textrm{append}(\textrm{NA})$ \COMMENT{either answer for $x_{ij} \leq t$ is possible}
\ENDIF
\ENDFOR
\STATE $X_T.\textrm{append}(x_T)$
\STATE $y_T.\textrm{append}(\hat{y}_\ell)$
\ENDFOR
\STATE Return $X_T$, $y_T$
\end{algorithmic}
\end{algorithm}

\textbf{Modified STreeD.} Algorithm \ref{alg:streed-change} shows the key adjustment made to an optimal tree solver: when a query cannot be evaluated for some sample (i.e., the value for that query is NA), we send the sample down both paths of the tree instead of one, penalizing either subtree for misclassifying that sample. Adding this modification to STreeD produces an algorithm that finds a tree that minimizes an upper bound on errors for the worst-case completion of a dataset with missing values (Lemma \ref{lemma:tree-change}). In particular, this means that if the modified algorithm reports a tree with 0 misclassification error for some dataset, that tree is guaranteed to have 0 error on any completion of the dataset. 
\begin{algorithm}[h]
\caption{\textsc{ConservativePartition}($\mathcal{D}, j$)}
\label{alg:streed-change}
\begin{algorithmic}[1]
\REQUIRE Dataset $\mathcal D$ where each feature has values $\in \{0, 1, \textrm{NA}\}$, feature $j$
\ENSURE Resulting partitions $\mathcal D_\textrm{left}, \mathcal D_\textrm{right}$, for which to evaluate accuracy and optimize subtrees.
\STATE {$\mathcal D_\textrm{left\phantom{1}} \gets \{(x_i, y_i) \in \mathcal D \mid x_{ij} = 1\} \cup \textcolor{red}{\{(x_i, y_i) \in \mathcal D \mid x_{ij} = \textrm{NA} \}}$}
\STATE $\mathcal D_\textrm{right} \gets \{(x_i, y_i) \in \mathcal D \mid x_{ij} = 0\} \cup \textcolor{red}{\{(x_i, y_i) \in \mathcal D \mid x_{ij} = \textrm{NA} \}}$
\STATE Return $\mathcal D_\textrm{left}, \mathcal D_\textrm{right}$
\end{algorithmic}
\end{algorithm}

\begin{restatable}{lemma}{TreeChange}\label{lemma:tree-change}
    Let $T^{mod}_{\mathcal D, d, \lambda}$ be a tree output from an optimal decision tree solver modified with Algorithm \ref{alg:streed-change}. Instead of being optimal for objective (\ref{eq:optimal-tree-objective}), this tree is optimal for an upper bound on Equation (1) for the worst-case completion of a dataset, subject to a sparsity penalty:
    $$T^{mod}_{\mathcal D, d, \lambda} \in \argmin_{T \in \mathcal T_d} \left(\lambda |T| + \sum_{\ell \in leaves(T)}\max_{\mathcal \{(x_{i'}, y_{i'})\}_{{i'}=1}^{|\mathcal D|}\in Completions(\mathcal D)} \sum_{i' =1}^{|\mathcal D|} \mathds{1}\{x_{i'} \in S(\ell) \wedge y_{i'} \neq \hat{y}_\ell\}\right)$$
\end{restatable}

Taken together, Lemmas \ref{remark:leaf-dataset} and \ref{lemma:tree-change} tell us that we can compose the modified solver and the modified dataset to find a sparsest form (Theorem \ref{thm:correct-sparsest}). Finding sparsest forms allows us to faithfully reduce ensembles that use binary leaf predictions in their probability estimates and predictions (Remark \ref{remark:ensprob}). One case of such ensembles are those found by the Adaboost algorithm \citep{freund1997decision} as implemented in scikit-learn \citep{scikit-learn}, which we explore in Section \ref{sec:experiments}.

\begin{restatable}{theorem}{CorrectSparsest}\label{thm:correct-sparsest}
    Under the same conditions as Lemma \ref{remark:leaf-dataset} and assuming there are no deterministic between-feature dependencies (e.g., two separate features representing age in months and age in years), Algorithm \ref{alg:sparsest} returns a sparsest decision equivalent form of the input tree within the specified depth bound, if one exists.
\end{restatable}

\begin{restatable}{remark}{ensprob}\label{remark:ensprob}
    Consider any ensemble of $m$ binary leaf decision trees  $C = [T^{(1)}, \ldots T^{(m)}] \in \mathcal T_d^m$ and weights $[w^{(1)}, \ldots, w^{(m)}]$. If the ensemble's probability prediction is 
    $$\hat{p}(x) = f\left(\frac{1}{\sum_{i=1}^m w^{(i)}}\sum_{i=1}^mw^{(i)}T^{(i)}(x)\right),$$
    for some function f, then its predicted probabilities are exactly equivalent for the ensemble 
    $[T'^{(1)}, \ldots T'^{(k)}] = \textsc{Reduced}(C)$, with weights $[w'^{(1)}, \ldots, w'^{(k)}]$ such that $w'^{(j)} = \sum_{i = 1}^m w^{(i)}\mathds{1}\{T^{(i)} \DE T'^{(j)}\} $.
\end{restatable}

\begin{algorithm}
    \caption{SparsestFormSolver(T, d)}
    \begin{algorithmic}[1]\label{alg:sparsest}
    \REQUIRE Tree $T$, search depth $d$
    \STATE $\textsc{STreeD}_\textrm{conservative-partition} \gets $ \textsc{STreeD} modified to use \textsc{ConservativePartition} (Algorithm \ref{alg:streed-change}) in place of the normal partition algorithm 
    \STATE $\mathcal D_{sparse} \gets \textsc{LeafDataset}(T)$ \COMMENT{Algorithm \ref{alg:make-dataset}}
    \STATE solver $\gets \textsc{STreeD}_\textrm{conservative-partition}$(depth $= d$, regularization $= \frac{1}{|T|^2 + 1}$)
    \STATE solver.fit($\mathcal D_{sparse}$)
    \STATE \textrm{Return} solver.tree
    \end{algorithmic}
\end{algorithm}

%% file: sections/experiments.tex
\subsection{Rashomon set Compression}

\begin{figure}[H]
    \centering
    \includegraphics[width=\linewidth]{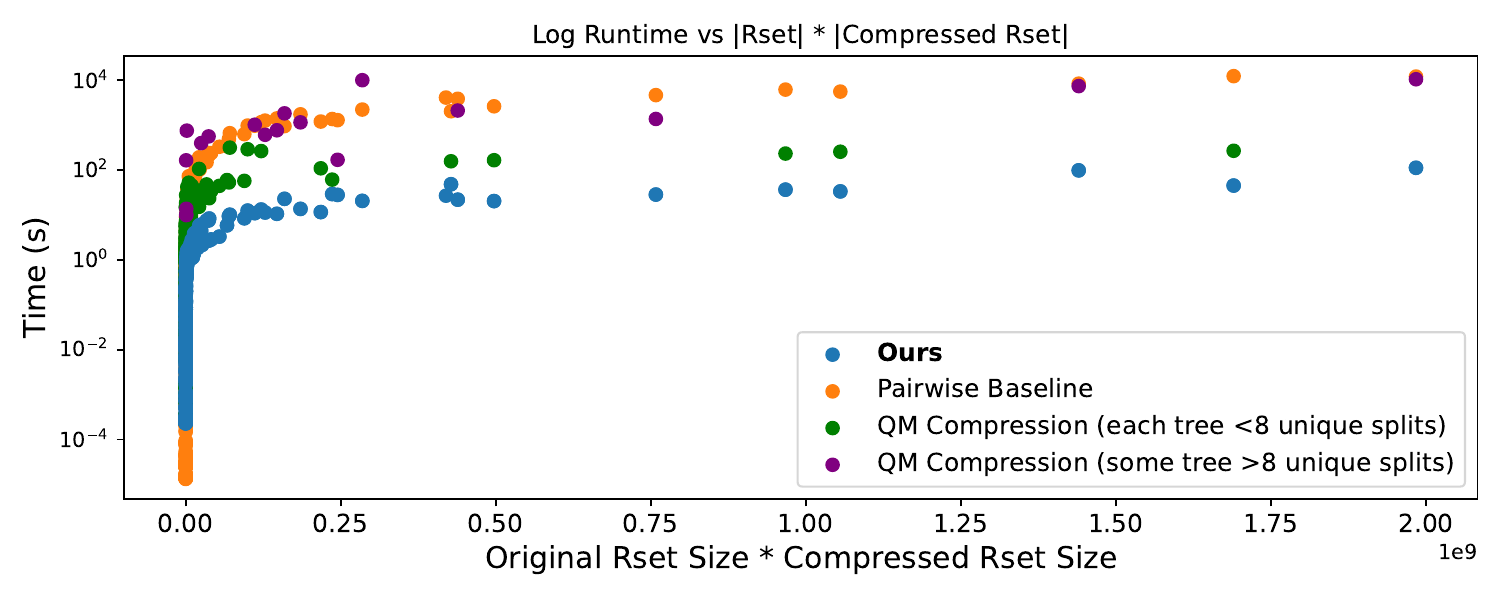}
    \caption{Time to simplify the Rashomon set, using Algorithm \ref{alg:collection} (Ours), with pairwise equivalence checks (Pairwise Baseline), or with logically simplifying each tree (QM Compression \cite{mctavish2025leveraging}). Results aggregated for all datasets in our suite, and all Rashomon sets of depth bound $[4, 5]$ (corresponding to maximum depths of 3 and 4), epsilon $[0.02, 0.05]$, regularization $[0.01, 0.005]$ that terminate within 24 hours and with 100GB of memory allowed. Our approach is several orders of magnitude faster, particularly for large, diverse Rashomon sets with trees that reason on more than a few splits.
    }\label{fig:rset-compress}
\end{figure}

Figure \ref{fig:rset-compress} demonstrates the scalability of our Rashomon compression (Section \ref{sec:pairwise}) relative to the baselines suggested by \cite{mctavish2025leveraging, marquessilva2025efficientpredictive}. 
We see that our approach scales substantially better in both the complexity of trees and the size of the set. We can handle complex individual trees much better than \cite{mctavish2025leveraging}, and we can handle larger Rashomon sets more scalably than \cite{marquessilva2025efficientpredictive}. We also demonstrate in Table \ref{tab:rset_size} that accounting for continuous feature dependencies further shrinks the Rashomon set. Revisiting datasets and configurations from \cite{mctavish2025leveraging} that use continuous features, some (though not all) datasets obtain 2-3x more compression of provably equivalent trees, undetectable to previous methods. 
These results demonstrate that decision equivalence in the Rashomon set is even more pervasive than previously thought and that our method can scalably \textbf{compress the Rashomon set by an order of magnitude}.

\begin{table*}[ht]
    \centering
    \begin{tabular}{c||c|c| c| c}
        Dataset & Total Trees & w/o Trivial \cite{xin2022exploring} & Boolean DE \cite{mctavish2025leveraging} & Ours\\
         \hline
         COMPAS & $12785 \pm 2840$ & $3913 \pm 837$ & $2135 \pm 448$ & $641 \pm 128$ \\
         Wine Quality & $6936 \pm 700$ & $2341 \pm 377$ & $1409 \pm 256$ & $436 \pm 90$ \\
         Wisconsin & $24052 \pm 9278$ & $11990 \pm 5283$ & $4657 \pm 1893$ & $4267 \pm 1728$\\
         FICO & $8086 \pm 2202$ & $1969 \pm 643$ & $1154 \pm 375$ & $752 \pm 254$ \\
    \end{tabular}
    \caption{Total \# of trees in the Rashomon set (maximum depth 3, epsilon 0.01), and \# of trees with methods to address equivalence: removing trivial redundancies (a filter provided in the original tree Rashomon set paper \cite{xin2022exploring}), removing decision equivalent trees assuming all boolean feature assignments are possible \cite{mctavish2025leveraging}, and removing decision equivalent trees accounting for continuous features (ours). Averaged over 5 folds.}
    \label{tab:rset_size}
\end{table*}

\subsection{Adaboost Ensemble Compression}
As a case study for sparsest form compression, we explore reducing the representation of Adaboost \citep{freund1997decision} classifiers, as implemented in scikit-learn. Concretely, this means identifying all trees in the same decision equivalence class and replacing them with a single tree that is in sparsest form, and has weight equal to the sum of weights of all the trees in its equivalence class. As discussed in Remark \ref{remark:ensprob}, this exactly preserves calibration of the original model. We demonstrate this reduction results in substantial compression of the classifier and
that the algorithm is practically feasible for shallow ensembles.

\textbf{Compression. }
Table \ref{tab:compression} shows the compression achieved between the reduced form and the original model. Because each tree is trained greedily, it is often quite far from the sparsest decision equivalent form. 
Occasionally, there are also multiple decision equivalent classifiers in the ensemble (these do not always occur but are sometimes significant), which can have their weight shifted to just one member of the equivalence class without changing the classifier's performance. 

\begin{table}[h]
    \centering
\begin{tabular}{lcccc}
\toprule
Dataset & Compression ratio & Proportion of Unique Trees \\
\hline
Iris\_setosa & 1.000 $\pm$ 0.000 & 1.000 $\pm$ 0.000 \\
Iris\_virginica & 0.740 $\pm$ 0.155 & 0.810 $\pm$ 0.163 \\
Iris\_versicolor & 0.727 $\pm$ 0.109 & 0.828 $\pm$ 0.127 \\
Spiral & 0.848 $\pm$ 0.009 & 0.958 $\pm$ 0.010 \\
Tic-Tac-Toe & 0.821 $\pm$ 0.005 & 1.000 $\pm$ 0.000 \\
Broward & 0.743 $\pm$ 0.024 & 0.852 $\pm$ 0.033 \\
COMPAS & 0.196 $\pm$ 0.048 & 0.246 $\pm$ 0.057 \\
Wisconsin & 0.972 $\pm$ 0.002 & 1.000 $\pm$ 0.000 \\
FICO & 0.776 $\pm$ 0.030 & 0.878 $\pm$ 0.031 \\
Wine Quality & 0.790 $\pm$ 0.016 & 0.942 $\pm$ 0.011 \\
Coupon & 0.764 $\pm$ 0.005 & 0.998 $\pm$ 0.002 \\
Netherlands & 0.441 $\pm$ 0.060 & 0.504 $\pm$ 0.085 \\
\end{tabular}
    \caption{Compression ratio ($\frac{\text{compressed ensemble leaf count}}{\text{original ensemble leaf count}}$) of an ensemble of 100 depth 3 estimators learned by Adaboost. Error bars are standard error from 5 folds. The full compression (both Sections \ref{sec:pairwise} and \ref{sec:sparsest}) is above 15 \% for most datasets. Even removing only equivalent pairwise forms without using the sparsest form finder results in some reduction in the number of trees.}
    \label{tab:compression}
\end{table}

There are two cases where compression is minor relative to the other datasets. Iris\_setosa, the only dataset without compression, is a very easy one-vs-rest problem where perfect accuracy is achieved by Adaboost with an early exit on a single decision stump. Note that nontrivial decision stumps are already in sparsest form. For Wisconsin, the other dataset without substantial compression (no duplicate trees and a compression ratio close to 1), many relevant feature dependencies exist that are out of scope for this work. Important features in this dataset, like the area, radius, and perimeter of a cell nucleus, clearly have dependencies that would warrant some dataset-specific modeling. This paper will miss out on compression where trees have distinct behaviour only in impossible or nearly impossible regions that depend on multiple different features, such as when area is high but perimeter is low. While we chose to focus in this work on only broader dependencies that can be modeled generally, exactly, and scalably across datasets (such as age <30 implying age <40), there is a clear direction for future work to model these additional dependencies approximately or exactly.

\textbf{Runtime.} Table \ref{tab:runtime} shows that our method efficiently compresses the trees in our ensembles. 
Most significantly, \textbf{we can frequently reduce Adaboost's complexity by >15\% in under a second}, reducing it to the reduced form defined in Definition \ref{def:reduced-form}. 

In the appendix, we show that using 100 depth 3 classifiers is comparable to the best tuned selection of hyperparameters for these datasets.

\begin{table}[h]
    \centering    
\begin{tabular}{lcc}
\toprule
Dataset & \textsc{Reduced} (Sections \ref{sec:pairwise} \& \ref{sec:sparsest}) & Remove DE trees (Section \ref{sec:pairwise}) 
\\
\hline
Iris\_setosa & 0.001 $\pm$ 0.000 & 0.000 $\pm$ 0.000 \\
Iris\_virginica & 0.161 $\pm$ 0.066 & 0.012 $\pm$ 0.005 \\
Iris\_versicolor & 0.254 $\pm$ 0.064 & 0.038 $\pm$ 0.024 \\
Spiral & 0.317 $\pm$ 0.023 & 0.018 $\pm$ 0.000 \\
Tic-Tac-Toe & 0.529 $\pm$ 0.020 & 0.023 $\pm$ 0.000 \\
Broward & 0.323 $\pm$ 0.027 & 0.025 $\pm$ 0.001 \\
COMPAS & 0.066 $\pm$ 0.010 & 0.018 $\pm$ 0.001 \\
Wisconsin & 0.789 $\pm$ 0.033 & 0.025 $\pm$ 0.001 \\
FICO & 0.422 $\pm$ 0.055 & 0.024 $\pm$ 0.001 \\
Wine Quality & 0.483 $\pm$ 0.025 & 0.048 $\pm$ 0.026 \\
Coupon & 0.557 $\pm$ 0.031 & 0.032 $\pm$ 0.000 \\
Netherlands & 0.215 $\pm$ 0.038 & 0.021 $\pm$ 0.002 \\

\end{tabular}
\caption{Runtime in seconds for compressing an ensemble of 100 depth 3 estimators learned by Adaboost. Error bars are standard error from 5 folds. The whole compression takes under a second.}
\label{tab:runtime}
\end{table}

%% file: sections/conclusion.tex
We have presented algorithms for losslessly compressing a collection of decision trees for binary classification. When there are multiple trees in the collection that make equivalent classifications for any possible input, we preserve only a single, sparsest tree from this equivalence class. 
We demonstrate this compression can be done scalably and completely for Rashomon sets and that we can compress ensembles while preserving their exact calibration for any possible sample. Future work could extend this compression to cases beyond binary classification trees, such as ensembles of trees with continuous leaf values \cite{chen2016xgboost} (one approximate way is to relax the decision equivalence check in Algorithm \ref{alg:de} to instead verify that two trees' predictions cannot deviate by more than some small amount).
Further efficiency improvements in reducing trees to their sparsest forms could allow this approach to scale to deeper ensembles. It may also be of interest to combine this work with other compression methods: Algorithm \ref{alg:collection} can be run as a quick preprocessing step to simplify a collection exactly before running further compression methods; and Algorithm \ref{alg:sparsest} can be run on each remaining tree in the output of a compressed ensemble to convert each tree to its sparsest equivalent form. 
The code used for experiments is available
at \url{https://github.com/HaydenMcT/decision-equivalent-tree-compression}.

%% file: sections/appendix/algorithms.tex
\section{Additional Algorithm Details}
\subsection{Decision Equivalence Solver}
Our decision equivalence algorithm, Algorithm \ref{alg:de}, works with a recursive subroutine over subregions, Algorithm \ref{alg:de-subroutine}. The structure is similar to \cite{zantema1998decision}, with changes made to remove some other runtime terms aside from the core $\mathcal O (|T||T'|)$ asymptotic bound. It works seamlessly with subregion tracking. 

\begin{algorithm}[h]
\caption{\textsc{FindNonEquivalence}(T,T')}
\begin{algorithmic}\label{alg:de}
\item[] {\bfseries Input:}  Decision trees $T, T'$. 
\item[] {\bfseries Output:} If $T(x) \DE T'(x)$, return None. Else, return a sample $x$ for which $T(x) \neq T'(x)$.
\item[] (Let $p$ denote dimensionality of domain of $T$)
\STATE open\_region $\gets (-\infty, \infty)^p$ \COMMENT{Initialize a data structure to track subregion contradictions (initializing with the open interval on all dimensions)}
\STATE return \textsc{SubregionContradiction}(open\_region, T, T') \COMMENT{Core logic is in Algorithm \ref{alg:de-subroutine}}
\end{algorithmic}
\end{algorithm}

\begin{algorithm}[h]
\caption{\textsc{SubregionContradiction}(S, T, T')}
\begin{algorithmic}[1]\label{alg:de-subroutine}
\item[] {\bfseries Input:}  Subregion S, decision trees $T$, $T'$. 
\item[] {\bfseries Output:} If $\forall x \in S, T(x) = T'(x)$, return None. Else, return a sample $x \in S$ for which $T(x) \neq T'(x)$.
\item[] {\bfseries Invariant:} $S$ is modified during execution, but will be in its original state at the end of execution
\IF {$T$ is a leaf}
\IF {$T'$ is a leaf}
\IF {$\hat{y}_{T} \neq \hat{y}_{T'}$}
\STATE return a sample in $S$
\ELSE
\STATE return None
\ENDIF
   \ELSE 
   \STATE return \textsc{SubregionContradiction}(S, $T'$, $T$) \COMMENT{\textbf{Note the swapped order of $T'$ and $T$}- we now iterate over $T'$, since $T$ has been reduced to a leaf}
   \ENDIF
   \ELSE
   \STATE $j, t$ = $T$.root.query \COMMENT{check query at $T$'s root}
   \IF{$(-\infty, t] \cap S_j \neq \emptyset$}
   \STATE old $\gets S_j$
   \STATE $S_j \gets S_j \cap (-\infty, t]$
   \STATE contradiction = \textsc{SubregionContradiction}(S, $T$.left, $T'$)
   \STATE $S_j \gets $ old \COMMENT{Restore original $S_j$ value}
   \IF{contradiction $\neq$ None}
   \STATE return contradiction
   \ENDIF
   \ENDIF
    \IF{$(t, \infty) \cap S_j \neq \emptyset$}
    \STATE old $\gets S_j$
   \STATE $S_j \gets S_j \cap (t, \infty)$
   \STATE contradiction = \textsc{SubregionContradiction}(S, $T$.right, $T'$)
   \STATE $S_j \gets $ old
   \IF{contradiction $\neq$ None}
   \STATE return contradiction   
   \ENDIF
   \ENDIF
   \ENDIF
\STATE Return None
\end{algorithmic}
\end{algorithm}

To check pairwise equivalence for baselines, we also use this subroutine (but a variant that does not generate a sample). This variant removes the $\mathcal O (p) $ runtime addition (which was only needed for sampling) and runs in $\mathcal O(|T||T'|)$ time. 

\subsection{Categorical Features}
The subregions we discuss assume continuous or binary features. Categorical features can be handled similarly for pairwise checks, as queries on whether a single member of a categorical feature is present are only known to be logically impossible if a different query on the same categorical feature is already known to be True, or if every other query on the categorical feature is known to be False (so we just track a count of confirmed false categorical levels relative to all possible; and track if one is known to be True). We focus in the main paper on algorithm details for continuous features, but the extension to categorical is possible with the same algorithms for pairwise checks, we just update subregion tracking as discussed above. 

%% file: sections/appendix/theory.tex
\section{Theorem Proofs} \label{app:proofs}

In this section we show proofs of the claims from the main paper, as well as a few supplementary Lemmas that are used to prove these claims.

\begin{lemma}\label{de-runtime}
For trees $T, T'$ operating over data with $p$ features, Algorithm \ref{alg:de} is correct and has runtime $\mathcal O(|T||T'| + p)$. 
\end{lemma}
\begin{proof}
First note that we always make at most one $\mathcal O(p)$ call (line 4 in Algorithm \ref{alg:de-subroutine}), which only occurs when we have decided to recurse no further because the trees are proven nonequivalent. It is sufficient to show that the decision version of the algorithm (every operation done other than sample generation) runs in $\mathcal O(|T||T'|)$ time; the sample generation is  then just a single extra $\mathcal O (p)$ operation. 

\textbf{Correctness:}

The proof will proceed by induction. We will show the subroutine Algorithm \ref{alg:de-subroutine} correctly evaluates whether $\forall x \in S, T(x) = T'(x)$ (and then it follows trivially that the call on an open interval from Algorithm \ref{alg:de} correctly maps to deciding decision equivalence). We also show the invariant holds ($S$ is the same at the start and end of recursive calls in Alg \ref{alg:de-subroutine}) . 

\textit{Base case: } $T, T'$ are both leaves. This part of the subroutine is clearly correct; decision equivalence is an easy problem for this region and we correctly return a sample in the subregion $S$ if the leaves are not equivalent, else None. The invariant on S is maintained because $S$ is not modified.

\textit{Inductive hypothesis: }  Let $T, T'$ be two trees and $S$ be the current subregion; assume Algorithm \ref{alg:de-subroutine} correctly evaluates decision equivalence over all subtrees of $T$ and $T'$ within their corresponding subregions.

\textit{Inductive step: } at least one of $T, T'$ is not a leaf. 
(Note that if $T'$ is not a leaf and $T$ is, we swap the arguments and continue, with no additional overhead and no further swaps. So we focus here on the case where $T$ is not a leaf). We make two recursive calls on smaller subproblems, checking if for all x in the subregion where the left path is taken, $T.left(x) = T'(x)$, and if for all x in the subregion where the right path is taken, $T.right(x) = T'(x)$. Since these cases partition the subregion, a positive answer in both cases (by the inductive hypothesis) guarantees the trees are decision equivalent for the current subregion. If either subroutine returns a sample, we propagate that sample correctly (since the sample is also inside this subregion if it was inside the subregion for one of the recursive calls). Note that we restore $S_j$ immediately after the recursive call where we modified the subregion, reversing the operation and restoring the original $S_j$.

Thus, by induction, the approach is correct.

The runtime is more immediate. We do a standard tree traversal through all leaves of $T$; for each leaf we do a tree traversal of tree $T'$. At each split we do a constant number of operations. Thus the runtime is $O(|T| |T'|)$, the product of the number of leaves in both trees. 

\end{proof}

\begin{remark}
    A variant of Algorithm \ref{alg:de} which returns True for nonequivalence and False for equivalence, without returning a sample, can achieve $\mathcal O (|T||T'|)$ complexity.
\end{remark}
\begin{proof}
    From Lemma \ref{de-runtime}, we know the subroutine runs in $\mathcal O(|T||T'|)$ except for generating the sample. Here we note simply that the initialization of the subregion can be done in $\mathcal O (1)$ time, if we represent the subregion valid intervals with a dictionary/hashmap, and note open intervals as corresponding to a key not existing in the dictionary. 
\end{proof}

\collectioncorrectness*
\begin{proof}
\phantom{newline}

\textbf{Runtime}

    We will argue that for each tree in $C_d$, we run a number of tree inferences and $\textsc{Pred\_Map}$ lookups proportional to the size of $X^{(\text{diff})}$ at that iteration, and will run at most one pairwise tree comparison. This will give us the corresponding runtime, for $\textsc{Pred\_Map}$ implemented as a hash table or similar data structure with fast lookups, and noting $|X^{(\text{diff})}| \leq |C'_d|$ throughout the algorithm.

    For each tree in $C_d$, we need to know if any prefix of the tree's predictions on $X^{(\text{diff})}$ is in $\textsc{Pred\_Map}$ (line 4); we can accomplish this by predicting with that tree on each sample in $X^{(\text{diff})}$ in sequence, checking each time for a collision with $\textsc{Pred\_Map}$ until either that collision occurs at index $k$ (making line 4 true) or until we have reached the end of $X^{(\text{diff})}$ with no collisions. In the latter case, we go to line 19, add in the new entry to $\textsc{Pred\_Map}$, and move on to the next tree, having done only $|X^{(\text{diff})}|$ inferences on the current tree and that many lookups on $\textsc{Pred\_Map}$. 

    Now consider the case where line 4 is True. Line 5's value has already been computed by the above process. Now we consider the tree $T'$ which has the same index in $\textsc{Pred\_Map}$ (line 6). To evaluate line 7, we predict with each of the two trees in sequence on $X^{(\text{diff})}$, starting from the (k + 1)$^{th}$ entry in $X^{(\text{diff})}$, and checking if the predictions ever differ before hitting the end of $X^{(\text{diff})}$. If they do, we can safely update $\textsc{Pred\_Map}$ with the two sets of predictions up through this index; in this case we then move on to the next tree in $C_d$, having in total done no more than 2 tree inferences for each sample in $X^{(\text{diff})}$, along with no more than $X^{(\text{diff})}$ queries on $\textsc{Pred\_Map}$ from the previous step. If they do not collide, then we move on to the else block at line 10. 
    
    Consider the case where we have reached the else block on line 10. We know the predictions through all of $X^{(\text{diff})}$, having done no more than 2 tree inferences for each sample in $X^{(\text{diff})}$, along with no more than $X^{(\text{diff})}$ queries on $\textsc{Pred\_Map}$. We now need to run one instance of the \textsc{FindNonEquivalence} algorithm. After that we are either done after adding the colliding tree back into the \textsc{Pred\_Map} with its new index (if the algorithm returns None) or need to run one final inference for each of the two trees on this differing point, and insert both trees back into $\textsc{Pred\_Map}$. So in total, before moving to the next tree, we had no more than $|X^{(\text{diff})}|$ inferences (where the size of $X^{(\text{diff})}$ may have increased by 1) for up to two trees, as well as the runtime of \textsc{FindNonEquivalence} and no more than $|X^{(\text{diff})}|$ lookups into $\textsc{Pred\_Map}$.

    The worst case of all of the above scenarios is no worse than $|X^{(\text{diff})}|$ lookups into  $\textsc{Pred\_Map}$, each O(1) for a hash table, $|X^{(\text{diff})}|$ inferences for two trees, each bounded by $O(d)$ where $d$ is the maximum depth of the trees, and a call to \textsc{FindNonEquivalence} which takes $\mathcal O(|T||T'|) + p$ time per Lemma \ref{de-runtime}. Considering worst case across the whole outer loop, and noting that $|X^{(\text{diff})}| \leq |C'_d|$ at every step (see the correctness proof below), we have runtime

    $$\mathcal O \left(|C_d| \left(|C'_d|d + \max_{T \in C_d, T' \in C'_d} |T||T'| + p\right)\right).$$

\textbf{Correctness}

    To prove that our collection satisfies the property that all equivalence classes are represented but there are no duplicates, we can use induction over the construction of $\textsc{Pred\_Map}$ and $X^{(\text{diff})}$. We operate here over two key invariants: $\textsc{Pred\_Map}$ contains a single member of all observed equivalence classes, indexed by a prefix of predictions on $X^{(\text{diff})}$ that is unique from all other equivalence classes.

    Base case: when we have not yet observed any samples and both $\textsc{Pred\_Map}$ and $X^{(\text{diff})}$ are empty, all invariants hold (vacuously).

    Inductive step: If we never hit a collision, then the invariants on $\textsc{Pred\_Map}$ and $X^{(\text{diff})}$ ensure that we have found a new equivalence class, and adding the tree to $\textsc{Pred\_Map}$ indexed with the full prediction vector on $X^{(\text{diff})}$ maintains the invariants (since we haven't hit a collision, we know the prediction vector on $X^{(\text{diff})}$ is unique from any other members of $\textsc{Pred\_Map}$). If we do hit a collision, then we consider three cases that cover the space of all possibilities: two trees are decision equivalent, they are not decision equivalent and they differ on $X^{(\text{diff})}$, and they are not decision equivalent and they do not differ on $X^{(\text{diff})}$. First, if further predictions on $X^{(\text{diff})}$ for the new tree and the colliding tree eventually cause a difference, we correctly maintain both trees (since they are not equivalent) and index them with a new, unique prefix from each other, maintaining invariants. 
    Second, if the prediction vectors continue to match and we ultimately find the two trees are decision equivalent, we reinsert the old colliding tree with the new index (still maintaining the invariant about unique prefix indices) and correctly do not add the new tree, since it is not a new equivalence class. If the prediction vectors continue to match and the trees are not decision equivalent, then we add a new sample to the end of $X^{(\text{diff})}$ (not violating any previous invariants, since all prefixes of prediction vectors on the old $X^{(\text{diff})}$ are still valid), and this new sample (as per Lemma \ref{de-runtime}) corresponds to a sample for which the two trees' predictions differ, so we can update the two prediction vectors and correctly add them, maintaining all invariants. 
\end{proof}

\collectionRashomon*
\begin{proof}
    To show this, it is sufficient to show that the decision equivalent form of a tree with the lowest objective maps to its simplest decision equivalent form. (From there, the remainder of the claim is a trivial consequence of the definition of the Rashomon set; all trees of lower objective than a member of the Rashomon set are also within the set, for a given depth $d$). 

    Recall Objective \ref{eq:rset} used to define the Rashomon set: 

    $$\textrm{Obj}(T, \mathcal D, \lambda) = \frac{1}{n}\sum_{i=1}^n \mathds{1}\{T(x_i) \neq y_i\} + \lambda |T|$$

    Without loss of generality, select two trees $T, T' \in \mathcal T_d$, $T \DE T'$, $\textrm{Obj}(T', \mathcal D, \lambda) \leq \textrm{Obj}(T, \mathcal D, \lambda)$. Then we know that $|T'| \leq |T|$ because the misclassification errors of the two trees must be the same. More formally: 
    \begin{align*}
        \textrm{Obj}(T', \mathcal D, \lambda) &\leq \textrm{Obj}(T, \mathcal D, \lambda)\\
        \frac{1}{n}\sum_{i=1}^n \mathds{1}\{T'(x_i) \neq y_i\} + \lambda |T'| &\leq \frac{1}{n}\sum_{i=1}^n \mathds{1}\{T(x_i) \neq y_i\} + \lambda |T|\\
        \frac{1}{n}\sum_{i=1}^n \mathds{1}\{T'(x_i) \neq y_i\} - \frac{1}{n}\sum_{i=1}^n \mathds{1}\{T(x_i) \neq y_i\} + \lambda |T'| &\leq \lambda |T|\\
        \frac{1}{n}\sum_{i=1}^n \left(\mathds{1}\{T'(x_i) \neq y_i\} - \mathds{1}\{T(x_i) \neq y_i\}\right) + \lambda |T'| &\leq \lambda |T|\\
        \frac{1}{n}\sum_{i=1}^n \left(\mathds{1}\{T(x_i) \neq y_i\} - \mathds{1}\{T(x_i) \neq y_i\}\right) + \lambda |T'| &\leq \lambda |T| \hspace{0.5em}(
        \text{By decision equivalence})\\
        \lambda |T'| &\leq \lambda |T|\\
        |T'| &\leq |T|\\
    \end{align*}
    As required. 

\end{proof}

\LeafDataset*
\begin{proof}
We begin by showing that \textbf{0 error implies decision equivalence}. 

Note that for a dataset containing every possible point in the domain $\mathcal{X}$, with labels given by $T'$'s predictions, 0 error on this dataset would yield a decision equivalent tree $T$ by definition.

Since leaves partition the domain, any possible point in the domain corresponds to a completion of a row in $\mathcal D_{missing}$. 

Since a tree's leaves are implicants for that tree's prediction, the labels for any completion of a row in $\mathcal D_{missing}$ will be consistent with $T'$'s classification of that completion. This is a consequence of the definition of completions and the fact that each row maps to a leaf implicant for tree $T'$: sample $x \in \mathcal X$ corresponds to a completion of row $i$ of $\mathcal D_{missing}$ if and only if $x$ is routed to leaf $\ell_i$ in $leaves(T')$. 

Therefore, if some tree $T$ has 0 error over any possible completion of any row in $\mathcal D_{missing}$, it must match tree $T'$'s predictions over any possible point in the domain. Then these two trees must be decision equivalent by definition.

We now show that \textbf{decision equivalence implies 0 error}. 

For most completions of the leaf dataset, this follows immediately from the definition of decision equivalence. We know from the definition of decision equivalence that for all realizable samples in the domain $\mathcal X$, two decision equivalent trees will classify those samples the same way. And we know the labels correspond to the predictions of $T'$ for all samples, because of how we have defined the missingness dataset. 

It is possible, however, for there to be some completion of the dataset that cannot be realized by any sample in the domain $\mathcal X$ (see Definition \ref{def:impossible}). 
\begin{definition}[Inconsistent assignments]\label{def:inconsistent} 
    Consider a set of splits of the form $\mathds{1}[v_j \leq t]$.
    Define an inconsistent assignment as an assignment of values to two splits that no $v \in \mathcal X$ can simultaneously satisfy.
    That is, for two thresholds such that $t_1 < t_2,$ any assignment such that $\mathds{1}[v_j \leq t_1] = 1 \wedge \mathds{1}[v_j \leq t_2] = 0$. 
    A path that requires such an inconsistent assignment to traverse is an impossible path.
\end{definition}
\begin{definition}[Impossible path]\label{def:impossible} 
    A path in a tree that requires an inconsistent assignment to traverse is an impossible path.
\end{definition}
To complete our proof, we need to show that two decision equivalent trees $T$ and $T'$ are also equivalent on any row completions of $\mathcal D_{missing}$ that correspond to inconsistent assignments. We handle this case with the following lemma: 

\begin{lemma}\label{lemma:matching}
    Let $T'$ and $T$ be trees over the domain $\mathcal{X}$ such that neither has impossible paths. Define $\mathcal D_{missing}$ to be the output of Algorithm \ref{alg:make-dataset} for $T'$. For any completion of $\mathcal D_{missing}$ containing an inconsistent assignment, there exists some other, in-domain completion of the same row of $\mathcal D_{missing}$ that is classified identically by $T$.
\end{lemma}
\begin{proof}
    Consider, without loss of generality, any row $r$ of $\mathcal D_{missing}$. 
    First note a simple property of Algorithm \ref{alg:make-dataset}: any completion of $r$ must be consistent with all known query values in $r$. This follows from the processing in Algorithm \ref{alg:make-dataset}, which only maintains NA in a column corresponding to $\mathds{1}[v_j \leq t]$ if both true and false values are possible given the known query values in that row.

    Now, consider any completion of this row containing an inconsistent assignment, and consider the leaf it reaches in $T$ (call it $\ell_T$. We know the inconsistent assignment cannot involve values fixed for $r$, by the argument above. 
    And the inconsistent assignment cannot be solely among values necessary to reach this leaf $\ell_T$ in $T$ -- that is, there must be at least one split involved in the inconsistent assignment that is not queried along the path to leaf $\ell_T$ -- because $T$ contains no impossible paths. 
    Thus the inconsistencies must involve some queries that are not necessary for reaching leaf $\ell_T$ and are not fixed for $r$ (that is, they are $NA$ in $r$). Since these values are $NA$, all possible assignments of these values are valid completions consistent with known values in $r$. We also know at least one valid set of assignments of these values will be consistent with reaching the leaf $\ell_T$, since no path in $T$ is impossible. Thus, we can construct another completion of row $r$ that is consistent and reaches the same leaf.
\end{proof}

With this lemma, we can complete the proof. Decision equivalence means all in-domain samples are classified identically for $T$ and $T'$. Now, for any out-of-domain sample, it must be classified the same way by $T$ as an in-domain sample completing the same row by Lemma \ref{lemma:matching}. Since both are completions of the same row, $T$ must also classify them correctly. Thus, $T$ will achieve 0 error, completing the proof.

\end{proof}
\newpage

\begin{figure}[H]
    \centering
\includegraphics[width=\linewidth]{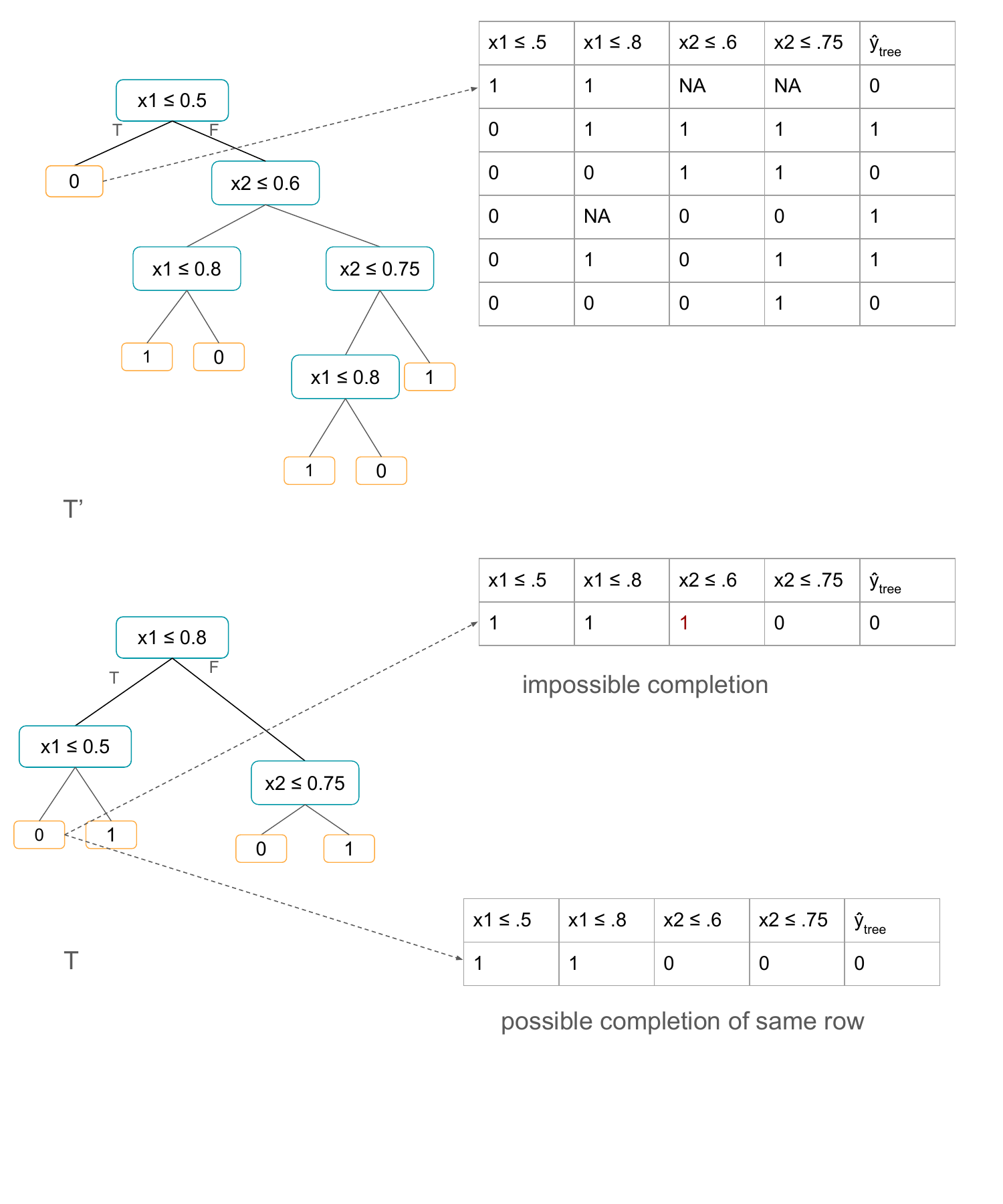}
\caption{Visualization of one example of matchings from lemma \ref{lemma:matching}.}
    \label{fig:leaf-dataset}
\end{figure}
\newpage

\TreeChange*

\begin{proof}
Note that recursive dynamic programming optimal tree solvers like STreeD work by recursively partitioning the data into left and right subregions, deciding on a tree making a split which corresponds to minimal error in these subregions (and filling in with optimal subtrees for these subregions); or making a leaf prediction if that is better.

The modification in Algorithm \ref{alg:streed-change} guarantees that as we consider splits, for each row in $\mathcal D$ where there exists a completion that is consistent with the subregion, the matched subregion will contain that row. This follows fairly directly: we can only remove a row from a subregion if it corresponds to the False (resp. True) path of a feature that is True (resp. False) and not NA (that is, the row is forced to conflict with that path no matter what). Otherwise, there exists at least one completion that is consistent with the path (either all completions are consistent, or we can fix an NA value to match that query). 

As such, we know that the number of errors a tree makes in any subregion where it predicts a leaf is at least the number of rows with completions that conflict with the leaf prediction. This follows because we include any row that has a conflicting label for at least one completion within that subregion. Assuming an optimizer that fixes $n$ in its objective calculation as the size of the original dataset (such an implementation detail is trivial to guarantee), that means minimizing misclassifications across all these subregions corresponds to minimizing an upper bound on the worst case completion of a dataset.
\end{proof}

\begin{remark}\label{remark:split-restriction}
For any tree $T'$ with splits that are of the form $\mathds{1}[x_{ij} \leq t]$, without deterministic between-feature dependencies in the feature space, there exists a sparsest equivalent form $T$ that uses a subset of the splits used in $T'$ and no other splits.
\end{remark}
\begin{proof}
Without loss of generality, let $\tilde{T}$ be a sparsest equivalent form of $T'$. Suppose it has some split $\mathds{1}[x_{ij'} \leq t']$ that does not occur in $T'$. Let $t^{above}$ and $t^{below}$ be the nearest thresholds in $T'$ on feature $j'$ above and below $t'$ (use $\infty$ or $-\infty$ if no such thresholds above or below exist, respectively). We will argue that each instance of the split $\mathds{1}[x_{ij'} \leq t']$ could be replaced with one of $\mathds{1}[x_{ij'} \leq t^{below}]$ or $\mathds{1}[x_{ij'} \leq t^{above}]$ while maintaining that the tree is equivalent to $T'$, thus forming a sparsest equivalent form using only splits from $T'$. 

Note that for tree $T'$, we are guaranteed to have the same prediction when we replace any sample that has a value for feature $j'$ in $(t^{below}, t^{above}]$ with another sample that has all other features kept the same, but feature $j'$ changed to any other value in $(t^{below}, t^{above}]$. All splits in tree $T'$ will be the same for these samples, so the path followed and the resulting prediction must be the same.

Since tree $\tilde{T}$ is decision equivalent to tree $T'$, it too must be guaranteed to have the same prediction when we replace any sample that has a value for feature $j'$ in $(t^{below}, t^{above}]$ with another sample that has all other features kept the same, but feature $j'$ changed to any other value in $(t^{below}, t^{above}]$. So any possible sample that has a value for feature $j'$ in $(t^{below}, t^{above}]$ and which reaches a split $\mathds{1}[x_{ij'} \leq t']$ must be able to go to either the left or right subtree of the split and receive the same prediction. These are also the only samples whose path could change if we replace the split $\mathds{1}[x_{ij'} \leq t']$ with either $\mathds{1}[x_{ij'} \leq t^{below}]$ or $\mathds{1}[x_{ij'} \leq t^{above}]$. Thus decision equivalence cannot be changed by doing such a substitution.

Thus, if there is ever a sparsest equivalent form that does not use only thresholds from $T'$, an equally sparse tree can be made that does use only thresholds from $T'$.
\end{proof}

\CorrectSparsest*
\begin{proof}

    This follows from Lemmas \ref{remark:leaf-dataset} and \ref{lemma:tree-change}. We do need a few extra details. First, note that we set the lambda value so low that the minima returned by Algorithm \ref{alg:sparsest} must make 0 errors. Because we have set lambda so low that its contribution to the objective on $T'$ is less than 1 misclassification, but the lambda is still nonzero, Algorithm \ref{alg:sparsest} must find the sparsest form achieving 0 error on this upper bound. 
    Second, note that 0 error for the summation in Lemma \ref{lemma:tree-change} can occur if and only if there is 0 error for the summation in Lemma \ref{remark:leaf-dataset}. If we had a nonzero error in either, it means we could find a nonzero error in the other. 
    
    Thus we have made a dataset such that 0 error across all completions occurs if and only if the found tree is decision equivalent to the original tree. 
    We then picked a hyperparameter setting for lambda such that this algorithm must achieve $0$ error in the sparsest form possible.

    Therefore, if a sparsest decision equivalent form exists, it will be returned.
\end{proof}

\ensprob*
\begin{proof}
For simplicity of notation, denote $\bar{w}^{(i)}:= \frac{w^{(i)}}{\sum_{i=1}^m w^{(i)}},$ giving
$$\hat{p}(x) = f\left(\sum_{i=1}^m\bar{w}^{(i)}T^{(i)}(x)\right)$$
Since we know that $C' \in  \textsc{Reduced}(C)$, from Definition \ref{def:reduced-form}, we know that each tree $T^{(i)} \in C$ is decision equivalent to exactly one tree $T'^{(j)}$ in the new collection. So, we have that:  
\begin{align*}
    \hat{p}(x) &=f\left(\sum_{i=1}^m\bar{w}^{(i)}T^{(i)}(x)\right)\\
    &=f\left(\sum_{i=1}^m\left(\bar{w}^{(i)}T^{(i)}(x)\sum_{j=1}^k\mathds{1}\{T^{(i)} \DE T'^{(j)}\}\right)\right) & \text{Multiply by 1}\\
    &=f\left(\sum_{j=1}^k\sum_{i=1}^m\left(\bar{w}^{(i)}T^{(i)}(x)\mathds{1}\{T^{(i)} \DE T'^{(j)}\}\right)\right) & \text{Swap summation order}\\
    &= f\left(\sum_{j=1}^k\sum_{i=1}^m\left(\bar{w}^{(i)}T'^{(j)}(x)\mathds{1}\{T^{(i)} \DE T'^{(j)}\}\right)\right) & \text{By definition of $\DE$}\\
    &= f\left(\sum_{j=1}^k\sum_{i=1}^m\left(\bar{w}^{(i)}\mathds{1}\{T^{(i)} \DE T'^{(j)}\}\right)T'^{(j)}(x)\right) \\
    &= f\left(\sum_{j=1}^k\bar{w}'^{(j)}T'^{(j)}(x)\right),
\end{align*}
where $\bar{w}'^{(j)}:=\sum_{i=1}^m\left(\bar{w}^{(i)}\mathds{1}\{T^{(i)} \DE T'^{(j)}\}\right)$. Thus, we have that $\hat{p}(x) = f\left(\sum_{j=1}^k\bar{w}'^{(j)}T'^{(j)}(x)\right)$ as required.
\end{proof}

%% file: sections/appendix/additional_results.tex
\section{Further Experimental Details}

\subsection{Additional Results}

Here we tune Adaboost on the following full grid using 5 fold cross validation. 'n\_estimators': [50, 100, 1000],'learning\_rate': [0.01, 0.1, 1], decision tree depth 1, 2, 3, 4, 5, no limit. We then compare to forcing 100 depth 3 estimators and only tuning the learning rate on the same values of [0.01, 0.1, 1]. Note that there is comparable AUC in both cases; we take this as evidence that the configuration restriction we use in the main paper is a reasonable restriction for AdaBoost. 

\begin{table}[ht]
    \centering
\begin{tabular}{lcc}
Dataset & AdaBoost & AdaBoost3 \\
\hline
Iris\_setosa & 1.000 $\pm$ 0.000 & 1.000 $\pm$ 0.000 \\
Iris\_virginica & 0.973 $\pm$ 0.011 & 0.960 $\pm$ 0.017 \\
Iris\_versicolor & 0.968 $\pm$ 0.018 & 0.989 $\pm$ 0.006 \\
Spiral & 0.691 $\pm$ 0.089 & 0.756 $\pm$ 0.062 \\
Tic-Tac-Toe & 0.998 $\pm$ 0.001 & 1.000 $\pm$ 0.000 \\
Broward & 0.650 $\pm$ 0.004 & 0.649 $\pm$ 0.006 \\
COMPAS & 0.734 $\pm$ 0.007 & 0.734 $\pm$ 0.007 \\
Wisconsin & 0.994 $\pm$ 0.003 & 0.994 $\pm$ 0.002 \\
FICO & 0.743 $\pm$ 0.011 & 0.745 $\pm$ 0.010 \\
Wine Quality & 0.819 $\pm$ 0.004 & 0.817 $\pm$ 0.005 \\
Coupon & 0.764 $\pm$ 0.004 & 0.760 $\pm$ 0.003 \\
Netherlands & 0.768 $\pm$ 0.002 & 0.764 $\pm$ 0.001 \\
\end{tabular}
    \caption{Test AUC when tuning a grid of parameter values for Adaboost, is comparable to that when constraining the grid to 100 depth 3 trees. Error bars are standard error for 5 outer folds (doing an inner 5 fold CV search for each outer fold). }
    \label{tab:accuracy}
\end{table}

\subsection{Computational Details}

Timing experiments for AdaBoost use Dell R730 Intel(R) Xeon(R) CPU E5-2640 v4 @ 2.4GHz,25M Cache (256GB RAM - 40 cores), with 50GB requested; other AdaBoost experiments use similar computing power. For the Rashomon set timing experiments with 100GB we use Exxact TS4-130227231 with AMD EPYC 9554 64-Core Processor (1TB RAM - 64 cores, 128 threads). Other experiments less sensitive to timing or hardware used whichever machines were available on our department cluster, with similar or less computational power. We suspect that most experiments except for the largest Rashomon sets run can be done quite feasibly with less; absolute timings will change but relative timing conclusions should be roughly consistent with the results from this paper. 

\subsection{Datasets and Assets}

We consider the 12 main datasets used in \cite{mctavish2025leveraging} which we refer to as Netherlands \cite{tollenaar2013method, DANS-23P-7UUU_2018}, Broward \cite{wang2023pursuit}, FICO \cite{competition}, Spiral \cite{mctavish2022fast}, Tic-Tac-Toe \cite{tictactoe}, and Iris Setosa/Versicolor/Virginica \cite{iris_53}, as well as COMPAS \cite{propublica}, Wine Quality \cite{cortez2009modeling}, Wisconsin \cite{street1993nuclear}, and Coupon \cite{wang2017bayesian}. COMPAS measures 7 features for 6,907 individuals, where labels are whether the individuals were arrested within 2 years of being released from prison. Wine Quality reports 11 features over 6,497 wines along with a numerical quality rating between 1 and 10. We binarize these ratings into high ($> 5$) and low ($\leq 5$) quality classes and predict this binary rating. Wisconsin is a breast cancer dataset and contains 30 features over 569 masses, where labels designate whether the tumor was malignant or benign. Coupon measures 25 features for 12,684 individuals, and labels denote whether or not the individual would accept a coupon.
Netherlands measures 9 features for 20,000 individuals, where labels are whether the individuals from the Netherlands committed another crime after being released from prison.
Similarly, Broward reports 38 features from 1,955 individuals from Broward county, Florida, where labels are whether the individuals from the committed another crime after being released from prison.
FICO contains 23 features from 10,459 individuals, and labels whether each individual will repay a line of credit within 2 years. In our experiments, we remove all rows from FICO that contain missing data, resulting in 2,502 samples.
Spiral is a synthetic dataset with 2 features over 100 samples, where each sample is randomly drawn from one of two interweaving spirals and the features are the x and y coordinates of the sample. In spiral, labels indicate which spiral each sample was drawn from.
Tic-Tac-Toe measures 9 features over all 958 possible games of tic-tac-toe, and labels each game as 1 if the first player won and 0 otherwise.
The three Iris datasets -- Iris Setosa, Iris Versicolor, and Iris Virginica -- are different tasks generated from the same multiclass classification dataset. The original Iris dataset reports 4 features over 150 iris flowers, and labels each sample as Setosa, Versicolor, or Virginica. We transform this dataset into three one versus all binary classification datasets.

Most datasets are from UCI, with a creative commons license. 
The license for FICO could previously be found at \url{https://community.fico.com/s/explainable-machine-learning-challenge?tabset-158d9=2}, although this link is no longer valid as of the camera-ready date. The license for Netherlands is available from contacting the authors of the paper \cite{tollenaar2013method}, with the dataset managed by DANS Data Station Physical and Technical Sciences (i.e., the managing entity for the formal dataset citation \cite{DANS-23P-7UUU_2018}). Spiral is a synthetic dataset from \cite{mctavish2022fast}.

We also make use of the codebase from \citet{mctavish2025leveraging} in our experimental comparisons and in some of the implementation of our own method; that codebase has an MIT license. 

\subsection{Preprocessing and further details}

To match prior work, in binarizing for Rashomon sets we used the internal binarization procedure from \cite{donnelly2023rashomon, mctavish2025leveraging}, which binarizes according to Threshold Guessing from \cite{mctavish2022fast}, using the thresholds selected based on 40 boosted decision stumps. We do not binarize for the Adaboost experiments, though we note for future work that quantizing could help with instances where there are two nearly identical splits in a tree. 

For quantifying the amount of compression our Rashomon sets match the parameters given in \cite{mctavish2025leveraging} for related experiments, and use depth 3 (i.e., the $depth\_bound$ parameter was set to 4), additive Rashomon bound 0.02, and sparsity penalty 0.01.

%% file: sections/appendix/scoping.tex
\section{Properties Preserved Within a Decision Equivalence Class}

This section includes a brief set of remarks about which metrics are preserved when we change a tree within its decision equivalence class. 

Broadly speaking, any property that depends solely on the classifier's prediction (on any possible sample, whether or not it is in-distribution) will be preserved. Properties that depend on the evaluation order implied by the tree, independently of the classifier's prediction, may not. 

For variable importance or feature attribution, remaining in a decision equivalence class for a single tree will not change any variable importance metric that obeys standard axioms \cite{lundberg2017unified, sundararajan2017axiomatic}. 
Some variable importance metrics, such as Gini importance, do not obey this property (as shown in \cite{mctavish2025leveraging}); we view this as a flaw in such variable importance metrics, rather than a limitation of changing a tree within an equivalence class. Another property that is not preserved is the presence of a meaningless feature - that is, a feature which, no matter what it is changed to, cannot influence predictions. All such splits are removed in the sparsest equivalent form of a tree.

In a setting where each feature maps to a cost (for example, different tests one might administer in a medical setting), selecting the sparsest equivalent form may not minimize the expected cost to evaluate the tree, or the most equitable partitioning of expected cost across subgroups.

We also note that one promising direction of future work -- to infer feature dependencies from data and define equivalence only within the set of points that are possible according to such a model -- can introduce complications that are not present in this work. Consider a biased dataset where some grouping of variables is a proxy for a protected attribute. In our method, we can never find a sparsest form that reasons directly on the protected attribute when given a tree that reasons only on the proxy. However, a method that infers feature dependencies may make this simplification based on the observed data. While we still view this direction of future work as promising, this risk would be a relevant consideration for such work.